\documentclass{article}
\usepackage{spconf,amsmath,graphicx,multirow,amssymb}
\usepackage{flushend}

\newcommand{\m}{\boldsymbol}
\newcommand{\argmin}{\mathop\textrm{argmin}}
\title{JOINT DICTIONARY LEARNING FOR EXAMPLE-BASED IMAGE SUPER-RESOLUTION}
\name{Mojtaba Sahraee-Ardakan$^1$, Mohsen Joneidi$^{2}$}
\address{$^1$Department of Electrical Engineering, Sharif University of Technology, Iran.\\
$^2$Department of Electrical Engineering, Ferdowsi University of Mashhad, Iran}
\begin{document}
%

\maketitle

\begin{abstract}
In this paper, we propose a new joint dictionary learning method for example-based image super-resolution (SR), using sparse representation. The low-resolution (LR) dictionary is trained from a set of LR sample image patches. Using the sparse representation coefficients of these LR patches over the LR dictionary, the high-resolution (HR) dictionary is trained by minimizing the reconstruction error of HR sample patches. The error criterion used here is the mean square error. In this way we guarantee that the HR patches have the same sparse representation over HR dictionary as the LR patches over the LR dictionary, and at the same time, these sparse representations can well reconstruct the HR patches. Simulation results show the effectiveness of our method compared to the state-of-art SR algorithms.
\end{abstract}
\section{Introduction}
\label{sec:intro}

Super-resolution is the problem of reconstructing a high resolution\footnote{In this article by resolution we mean spatial resolution.} image from one or several low resolution images \cite{park2003super}. It has many potential applications like enhancing the image quality of low-cost imaging sensors (e.g., cell phone cameras) and increasing the resolution of standard definition (SD) movies to display them on high definition (HD) TVs, to name a few.

Prior to SR methods, the usual way to increase resolution of images was to use simple interpolation-based methods such as bilinear, bicubic and more recently the resampling method described in \cite{zhang2006edge} among many others. However all these methods suffer from blurring high-frequency details of the image especially for large upscaling factors (the amount by which the resolution of image is increased in each dimension). Thus, over the last few years, a large number of SR algorithms have been proposed \cite{milanfar2010super}. These methods can be classified into two categories: multi-image SR, and single-image SR.

Since the seminal work by Tsai and Huang \cite{tsai1984multiframe} in 1984, many multi-image SR techniques were proposed \cite{elad1997restoration,elad1999super,capel2001super,farsiu2004fast}. In the conventional SR problem, multiple images of the same scene with subpixel motion are required to generate the HR image. However the performance of these SR methods are only acceptable for small upscaling factors (usually smaller than 2). As the upscaling factor increases, the SR problem becomes severely ill-conditioned and a large number of LR images are needed to recover the HR image with acceptable quality.

To address this problem, example-based SR techniques were developed which require only a single LR image as input \cite{freeman2002example}. In these methods, an external training database is used to learn the correspondence between manifolds of LR and HR image patches. In some approaches, instead of using an external database, the patches extracted from the LR image itself across different resolutions are used \cite{glasner2009super}. In \cite{freeman2002example} Freeman \textit{et al.} used a Markov network model for super-resolution. Inspired by the ideas in locally linear embedding (LLE) \cite{roweis2000nonlinear}, the authors of \cite{chang2004super} used the similarity between manifolds of HR patches and LR patches to estimate HR image patches. Motivated by results of compressive sensing \cite{candes2006compressive}, Yang \textit{et al.} in \cite{yang2008image} and \cite{yang2010image} used sparse representation for SR. In \cite{yang2012coupled} they introduced coupled dictionary training in which the sparse representation of LR image patches better reconstructs the HR patches.

Recently, joint and coupled learning methods are utilized for efficient modeling of correlated sparsity structures \cite{7301775,yang2010image}. However joint learning methods and the coupled learning methods proposed in \cite{yang2008image,yang2010image,7533090} still does not guarantee that the sparse representation of HR image patches over the HR dictionary is the same as the sparse representation of LR patches over LR dictionary. To address this problem, in this paper we propose a direct way to train the dictionaries that enforces the same sparse representation for LR and HR patches. Moreover since the HR dictionary is trained by minimizing the final error in reconstruction of HR patches, the reconstruction error in our method is smaller.

The rest of this paper is organized as follows. In section 2, Yang's method for super-resolution via sparse representation is reviewed. In section 3, a flaw in Yang's method is discussed, and our method to solve this problem is presented. Finally, section 4 is devoted to simulation results.
\section{review of super-resolution via sparse representation}
\label{sec:review}

In SR via sparse representation we are given two sets of training data: a set of LR image patches, and a set of corresponding HR image patches. In other words, in the training data we have pairs of LR and HR image patches. The goal of SR is to use this database to increase the resolution of a given LR image.

Let $\{\m{y}_i\}_{i=1}^{N}$ be the set of LR patches (each patch is arranged into a column vector $\m{y}_i$) and $\{\m{x}_i\}_{i=1}^N$ be the set of corresponding HR patches. In SR using sparse representation, the problem is to train two dictionaries $\m{D}_l$ and $\m{D}_h$ for the set of LR patches (or a feature of these patches) and HR patches respectively, such that for any LR patch $\m{y}_i$, its sparse representation $\m{w}_i$ over $\m{D}_l$, reconstructs the corresponding HR patch $\m{x}_i$ using $\m{D}_h$: $\m{x}_i\approx\m{D}_h\m{w}_i$ ~\cite{yang2010image}. Towards this end, first the dictionary learning problem is briefly reviewed in section ~\ref{ssec:dic}. Then the dictionary learning method for SR proposed in \cite{yang2010image} is studied in section ~\ref{sec:yangdic}. Finally in section ~\ref{sec:SR}, it is shown how these trained dictionaries can be used to perform SR on a LR image.

\subsection{Dictionary learning}
\label{ssec:dic}

Given a set of signals $\{\m{x}_i\}_{i=1}^N$, dictionary learning is the problem of finding a wide matrix $\m{D}$ over which the signals have sparse representation ~\cite{elad2010sparse}. This problem is highly related to subspace identification \cite{rahmani2015innovation}. However, sparsity helps us to turn the subspace recovery to a well-defined problem. This approach has attracted lot of attentions in the last decade and found diverse applications \cite{minaee2015screen,abavisani2015robust,joneidi2015eigen}. If we denote the sparse representation of $\m{x}_i$ over $\m{D}$ by $\m{w}_i$, the dictionary learning problem can be formulated as
\begin{equation}
\min_{\m{D},\{\m{w}_i\}_{i=1}^N}{\sum_{i=1}^N{\|\m{w}_i\|_0}}\quad s.t.   \|\m{x}_i-\m{D}\m{w}_i\|_2^2\leq\epsilon , i=1,...,N \label{diclearn}
\end{equation}
in which the $\|\cdot\|_0$ is the $l_0$-norm which is the number of nonzero components of a vector and $\epsilon$ is a small constant which determines the maximum tolerable error in sparse representations. Replacing the $l_0$-norm by $l_1$-norm, Yang \textit{et al.}\ in \cite{yang2010image} used the following formulation for sparse coding instead of \eqref{diclearn}
\begin{equation}
\min_{\m{D},\{\m{w}_i\}_{i=1}^N}{\sum_{i=1}^N{\|\m{x}_i-\m{D}\m{w}_i\|_2^2}+\lambda\sum_{i=1}^N{\|\m{w}_i\|_1}}.\label{yangdiclearn}
\end{equation}
By defining $\m{X}\triangleq [\m{x}_1 {\ } \cdots{\ } \m{x}_N]$ and $\m{W}\triangleq[\m{w}_1 {\ } \cdots{\ } \m{w}_N]$, it can be rewritten in matrix form as
\begin{equation}
\min_{\m{D},\m{W}}{\|\m{X}-\m{D}\m{W}\|_F^2+\lambda{\|\m{W}\|_1}},\label{yangdiclearn2}
\end{equation}
in which $\|\cdot\|_F$ stands for the Frobenius norm. \eqref{yangdiclearn} and \eqref{diclearn} are not equivalent, but closely related. \eqref{yangdiclearn} can be interpreted as minimizing the representation error of signals over the dictionary, while forcing these representations to be sparse by adding a $l_1$-regularization to the error. Therefore $\lambda$ can be used as a parameter that balances the sparsity and the error; a larger $\lambda$ results in sparser representations with larger errors.
\subsection{Dictionary learning for SR}
\label{sec:yangdic}
Given the sets of LR and HR training patches, $\{\m{y}_i\}_{i=1}^{N}$ and $\{\m{x}_i\}_{i=1}^{N}$, by defining $\m{Y}\triangleq[\m{y}_1 {\ } \cdots{\ } \m{y}_N]$ , and having \eqref{yangdiclearn2} in mind, Yang \textit{et al.}\ in \cite{yang2010image} proposed the following joint dictionary learning to ensure that the sparse representation of LR patches over $\m{D}_l$ is the same as sparse representation of HR image patches over $\m{D}_h$:
\begin{equation}
\min_{\m{D}_l,\m{D}_h,\m{W}}{{\|\m{Y}-\m{D}_l\m{W}\|_F^2}+{\|\m{X}-\m{D}_h\m{W}\|_F^2}+\lambda{\|\m{W}\|_1}}\label{yanggg}
\end{equation}
The key point here is that they have used the same matrix $\m{W}$ for sparse representation of both LR and HR patches to make sure that their representation is the same over the dictionaries $\m{D}_l$ and $\m{D}_h$. If we define the concatenated space of HR and LR patches:
\begin{equation*}
\m{Z}=\left[\begin{matrix}
\m{Y} \\ 
\m{X}
\end{matrix}\right],\quad \m{D}=\left[\begin{matrix}
\m{D}_l \\ 
\m{D}_h
\end{matrix}\right]
\end{equation*}
then joint dictionary training \eqref{yanggg} can also be written equivalently as
\begin{equation}
\min_{\m{D},\m{W}}{\|\m{Z}-\m{D}\m{W}\|_F^2+\lambda{\|\m{W}\|_1}}.\label{con}
\end{equation}
This formulation is clearly the same as \eqref{yangdiclearn2}. In other words, in the concatenated space, joint dictionary learning is the same as conventional dictionary learning, and any dictionary learning algorithm can be used for joint dictionary learning.
\subsection{Super-Resolution}
\label{sec:SR}
After training the two dictionaries $\m{D}_l$ and $\m{D}_h$, the input LR image can be super-resolved using the following steps:

\begin{enumerate}
\item The input LR image is divided into a set of overlapping LR patches: $\{\m{y}_i^{LR}\}_{i=1}^M$.
\item From each image patch $\m{y}_i^{LR}$, subtract its mean, $\eta_i$,
\begin{equation*}
{\hat{\m{y}}_i^{LR}}\triangleq\m{y}_i^{LR}-\eta_i,
\end{equation*}
and find its sparse representation over $\m{D}_l$
\begin{equation*}
\m{w}_i=\arg \min_{\m{\alpha}_i}{\|\hat{\m{y}}_i^{LR}-\m{D}_l\m{\alpha}_i\|_2^2+\lambda\|\m{\alpha}_i\|_1}.
\end{equation*} 
\item Using the sparse representation of each LR patch and its mean, the corresponding HR patch is estimated by
\begin{equation*}
\hat{\m{x}}_i^{HR}=\m{D}_h\cdot \m{w}_i,\quad {\m{x}}_i^{HR}=\hat{\m{x}}_i^{HR}+\eta_i.
\end{equation*}
\item Combining the estimated HR image patches, the output HR image is generated.
\end{enumerate}
\section{our proposed method}
\label{ours}

Our method for SR is to improve the dictionary learning part of Yang's method, described in section \ref{sec:yangdic}. Having the dictionaries trained, the rest of the method is the same as what described in section \ref{sec:SR}.

As mentioned earlier, in SR the dictionaries should be trained in a way that the sparse representation of each LR patch well reconstructs the corresponding HR patch. The Yang's method uses \eqref{yanggg} to accomplish this. It uses the same sparse representation matrix $\m{W}$ for both LR and HR patches to ensure that each LR and HR patch, both have the same sparse representation. However as can be seen from \eqref{con}, this joint dictionary learning is only optimal in the concatenated space of LR and HR patches, but if we look at the space of LR and HR patches separately, we may find a sparser representation for some patches than the sparse representation found in the concatenated space.

To address this problem, note first that the SR method described in section \ref{sec:SR} consists of two distinct operations: finding the sparse representation of the LR patch, and the reconstruction of the HR patch. Then, we note that the first operation uses only $\m{D}_l$, and the second operation uses only $\m{D}_h$. Therefore, instead of training the dictionaries jointly as in \eqref{yanggg}, we propose to train $\m{D}_l$ for LR patches solely, and then to train the HR dictionary by minimizing the reconstruction error when sparse representation of LR patches are used.

Mathematically, we propose to train the LR dictionary as
\begin{equation}
\min_{\m{D}_l,\m{W}}{\|\m{Y}-\m{D}_l\m{W}\|_F^2+\lambda{\|\m{W}\|_1}},\label{mylr}
\end{equation}
which is a conventional dictionary learning problem. After training the LR dictionary, for each LR patch, its sparse representation $\m{w}_i$ is found over $\m{D}_l$ (note that this step is already done during the dictionary training in \eqref{mylr})
\begin{equation}
\m{W}=\argmin_{\m{V}}{\|\m{Y}-\m{D}_l\m{V}\|_F^2+\lambda{\|\m{V}\|_1}}.
\end{equation}
Using the sparse representation of LR patches $\m{W}$, the HR dictionary $\m{D}_h$ is found such that the reconstruction error of the corresponding HR patches are minimized, that is,
\begin{equation}
\m{D}_h=\argmin_{\m{D}_h}{\|\m{X}-\m{D}_h\m{W}\|_F^2}.\label{HRtrain}
\end{equation}
This is an unconstrained quadratic optimization problem which has the following closed-form solution:
\begin{equation}
\m{D}_h=\m{X}\m{W}\left(\m{WW}^T\right)^{-1}=\m{X}\m{W}^\dagger
\end{equation}
in which $(\cdot)^T$ and $(\cdot)^\dagger$ represent transpose and pseudo-inverse of a matrix, respectively.

Note that unlike Yang's method, in the proposed method $\m{D}_h$ is not trained in a way that explicitly enforces the sparsity of representation of HR patches over it, rather it is trained to minimize the final reconstruction error.

\section{simulation results}
\label{sec:sim}
\begin{figure}[t]

\begin{minipage}[b]{0.5\linewidth}
  \centering
  \centerline{\includegraphics[width=4cm]{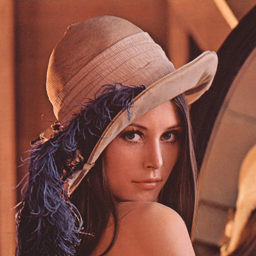}}
  \centerline{(a) Original}\medskip
\end{minipage}
\begin{minipage}[b]{0.45\linewidth}
  \centering
  \centerline{\includegraphics[width=4cm]{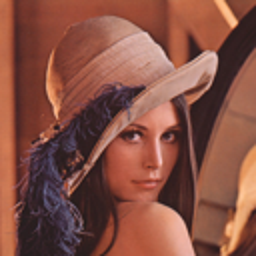}}
  \centerline{(b) Bicubic interpolation}\medskip
\end{minipage}

\begin{minipage}[b]{0.5\linewidth}
  \centering
  \centerline{\includegraphics[width=4cm]{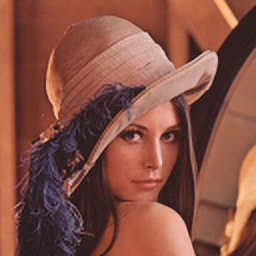}}
  \centerline{(c) Yang's method}\medskip
\end{minipage}
\begin{minipage}[b]{0.45\linewidth}
  \centering
  \centerline{\includegraphics[width=4cm]{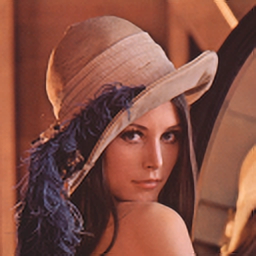}}
  \centerline{(d) Proposed method}\medskip
\end{minipage}
\caption{Results of Lena image magnified by a factor of 2 using: (b) Bicubic interpolation, (c) Yang's method, (d) our proposed method. The original image is also given in (a) for comparison. }
\label{fig:res}
\end{figure}

In this section we compare the performance of our method with Yang's method. The error criteria used here are {Peak Signal to Noise Ratio} (PSNR) and Structural SIMilarity (SSIM) index \cite{wang2004image}. PSNR criterion is defined as
\begin{equation}
\textrm{PSNR}=20\log_{10}\left(\frac{255}{\sqrt{MSE}}\right)\label{psnr},
\end{equation}
where MSE is mean square error given by
\begin{equation}
\textrm{MSE}=\frac{\|\m{I}_{SR}-\m{I}_g\|_F^2}{mn}\label{mse},
\end{equation}
in which  $\m{I}_g$ is the original distortion-free image, and $\m{I}_{SR}$ is the super-resolved image derived from the SR algorithm, and $m$ and $n$ are dimensions of the image in pixels.

For the definition of SSIM refer to \cite{wang2004image}. From these definitions it is clear that higher PSNR means less mean square error, however it does not necessarily mean a better image quality when perceived by human eye. Many other error criteria have been proposed to solve this problem of PSNR. SSIM is one of these error criteria. But still PSNR is widely used because of its simple mathematical form. This can be seen in \eqref{HRtrain} where MSE is to train the HR dictionary, but in order to show the effectiveness of our method, here we use both SSIM and PSNR to compare images produced by our method with Yang's method.

To make a fair comparison, the same set of 80000 training data patches sampled randomly from natural images is used to train dictionaries for both Yang's and our method. The size of LR patches is $5\times5$ and they are magnified by a factor of 2, i.e.\ the size of generated HR image patches is $10\times10$. The LR patches extracted from the input image have a 4 pixel overlap. Dictionary size is fixed at 1024, and $\lambda=0.15$ is used for both methods as in \cite{yang2010image}.

In Fig.\ \ref{fig:res}, simulation results of Yang's method and proposed method on Lena image can be seen. The original image and the image magnified using bicubic interpolation are also given as references. The PSNRs of these images are $32.79$dB, $34.73$dB and $34.86$dB for bicubic interpolation, Yang's method and ours respectively. It is clear that the quality of images magnified by SR is much better than the image magnified by bicubic interpolation and the details are more visible, which has resulted in sharper images. But the difference between image (c) and image (d) is not noticeable visually, although the PSNR of image (d) which is super-resolved by our method is about $0.1$dB higher.
{\renewcommand{\arraystretch}{1.3}
\renewcommand{\tabcolsep}{0.2cm}
\begin{table}[t!]
\centering
\caption{PSNR and SSIM of some images magnified using bicubic iterpolation, Yang's method and our proposed method. The average PSNRs and SSIMs are given in the last row. Best performance in each row is written in boldface.}\label{tab}
\begin{tabular}{cc|*{3}{c|}}
\cline{3-5}
& &{Bicubic} & {\ \ Yang\ \ } &{proposed} \\ 

\hline
\multicolumn{1}{|c|}{\multirow{2}{*}{Lena}} &PSNR& 32.79& 34.73& \bf{34.86}\\ 

\multicolumn{1}{ |c|  }{}&SSIM& 0.9012& 0.9268& \bf{0.9283}\\
\hline 
\multicolumn{1}{|c|}{\multirow{2}{*}{Parthenon}} &PSNR& 26.50 & 27.77& \bf{27.89} \\ 
\multicolumn{1}{ |c|  }{}&SSIM& 0.8334 & 0.8737& \bf{0.8762} \\
\hline 
\multicolumn{1}{|c|}{\multirow{2}{*}{Baboon}} &PSNR& 24.66& 25.30& \bf{25.39} \\ 
\multicolumn{1}{ |c|  }{}&SSIM& 0.9529& 0.9872& \bf{0.9873} \\
\hline 
\multicolumn{1}{|c|}{\multirow{2}{*}{Barbara}} &PSNR& 27.93 & \bf{28.61}& 28.59  \\ 
\multicolumn{1}{ |c|  }{}&SSIM& 0.9609 & \bf{0.9852}& \bf{0.9852}  \\ 
\hline 
\multicolumn{1}{|c|}{\multirow{2}{*}{Flower}} &PSNR& 30.51& 33.24 & \bf{33.36} \\ 
\multicolumn{1}{ |c|  }{}& SSIM&0.9230& 0.9526 & \bf{0.9538} \\
\hline \hline
\multicolumn{1}{|c|}{\multirow{2}{*}{\bf{Average}}} &PSNR& 28.47 & 29.93 & \bf{30.02} \\ 
\multicolumn{1}{ |c|  }{}&SSIM& 0.9143&0.9451&\bf{0.9462}\\
\hline 

\end{tabular} 

\end{table}}

In Table \ref{tab} the PSNRs and SSIMs of some images produced by our method is compared with those of Yang's method and bicubic interpolation. Almost all of the images recovered by our method have higher PSNRs than images recovered by Yang's method. The average PSNRs given in the last row show that our method performs slightly better than Yang's method on average.

The SSIMs in Table \ref{tab} also confirm that our method is performing better than Yang's method. The images super-resolved by the proposed method have on average a higher SSIM than images recovered by Yang's method. Since SSIM is much more consistent with the image quality as it is perceived  by human eye compared to PSNR, higher SSIM of images recovered by our method suggests that they also have better visual quality.

\section{Conclusion and Future Works}
\label{sec:conclusion}

In this paper, we presented a new dictionary learning algorithm for example-based SR. The dictionaries were trained from a set of sample LR and HR image patches in order to minimize the final reconstruction error. Simulation results on real images showed the effectiveness of our algorithm in super-resolving images with less error compared to Yang's method. The average PSNR and average SSIM of images produced by our method were higher than images recovered by Yang's method. In future, we can extend this work by training the HR dictionary using a better error criterion instead of PSNR. One of the advantages of our method is that training of $\m{D}_h$ is separated from $\m{D}_l$ in \eqref{mylr} and \eqref{HRtrain}. We can use another error criterion that better represents the image quality like SSIM in \eqref{HRtrain} without making the training of $\m{D}_l$ more complex. Changing the error criterion in each of Yang's methods will make the optimizations in their algorithms much more complex.




\bibliographystyle{IEEEbib}
\bibliography{strings,refs}

\end{document}